\RequirePackage{fix-cm}
\documentclass[smallextended]{svjour3}       
\smartqed  

\usepackage[namelimits]{amsmath}
\usepackage{graphicx}
\usepackage{booktabs}
\usepackage{float}

\usepackage{graphicx}
\usepackage{booktabs}
\usepackage{float}
\usepackage{subfig}
\usepackage{cite}
\usepackage{url}
\usepackage{amssymb}
\usepackage{bm}

\begin{document}

\title{Multi-feature Distance Metric Learning for Non-rigid 3D Shape Retrieval
}


\author{Huibing Wang         \and
        Haohao Li         \and
        Xianping Fu \footnote{Corresponding Author} 
}


\institute{H. Wang \at
              College of Information and Science Technology, Dalian Maritime University, Dalian, China, 116021\\
              \email{huibing.wang@dlmu.edu.cn}           
           \and
           H. Li \at
              School of Mathematical Sciences, Dalian University of Technology, Dalian, China, 116024\\
              \email{hhl820@mail.dlut.edu.cn} 
          \and
           X. Fu \at
              College of Information and Science Technology, Dalian Maritime University, Dalian, China, 116021\\
              \email{fxp@dlmu.edu.cn} 
}

\date{Received: date / Accepted: date}

\maketitle

\begin{abstract}
In the past decades, feature-learning-based 3D shape retrieval approaches have been received widespread attention in the computer graphic community. These approaches usually explored the hand-crafted distance metric or conventional distance metric learning methods to compute the similarity of the single feature. The single feature always contains onefold geometric information, which cannot characterize the 3D shapes well. Therefore, the multiple features should be used for the retrieval task to overcome the limitation of single feature and further improve the performance. However, most conventional distance metric learning methods fail to integrate the complementary information from multiple features to construct the distance metric. To address these issue, a novel multi-feature distance metric learning method for non-rigid 3D shape retrieval is presented in this study, which can make full use of the complimentary geometric information from multiple shape features by utilizing the KL-divergences. Minimizing KL-divergence between different metric of features and a common metric is a consistency constraints, which can lead the consistency shared latent feature space of the multiple features. We apply the proposed method to 3D model retrieval, and test our method on well known benchmark database. The results show that our method substantially outperforms the state-of-the-art non-rigid 3D shape retrieval methods.
\end{abstract}

\section{Introduction}
\label{intro}
With the development of information technology \cite{wu2019cycledeep,wu2018deepadatpt}, non-rigid 3D shape retrieval has been an active research spot for many years with explosive growth of 3D models \cite{reuter2006laplace,bronstein2011shape,lian2013comparison,litman2014supervised,xie2017deepshape,wu20183d}. The 3D shape retrieval are described as: Given a set of 3D shapes and a query shape, we would like to develop an effective algorithm to measure the similarity of the query \cite{wang2015effective} to all shapes in the datebase. 3D models have the complicated geometric structure information, which is difficult to construct the discriminative global features for various application. Only onefold global features usually cannot characterize the 3D shapes well, which means that the only onefold intrinsic geometric information is not enough to discriminate various 3D shapes for the non-rigid 3D shape retrieval task. Meanwhile the non-rigid deformations of the shapes induce the noise of the features, which impacts the computation of 3D shape similarity. Therefore, how to effectively calculate the distance between non-rigid 3D shapes is still a challenging problem.

In recent years, various non-rigid 3D shape retrieval algorithms had been proposed. Most of the algorithms focus on extracting the intrinsic features of the shapes based on the local or global geometric structure and measuring the similarity of the features. These approaches usually extract novel intrinsic features for the shapes firstly. Then, the hand-crafted distance metric or conventional distance metric learning methods are used to compute the similarity of the features. In \cite{reuter2006laplace}, the bag-of-word feature with spatial information are constructed by coding the spectral signatures. Then the Similarity Sensitive Hashing (SSH) are used to improve performance of the retrieval. Litman \cite{litman2014supervised} extract the global features by sparse dictionary learning algorithm, and explore the Euclidean metric to measure the similarity between the features. These methods explore single intrinsic features, which is not enough for discriminating various 3D shapes. Different with the single features, multiple features contain compatible and complementary geometric information which can improve the performance of the retrieval task. Chiotellis \cite{chiotellis2016non}, use the weighted averaging directly on two spectral signatures to construct the global features, and the similarity of the features are measured by Large margin nearest neighbor algorithm. Some approaches \cite{xie2017deep,ohbuchi2010distance,lian2011shape} explore weighted averaging of the distance between single feature to measure the similarity of the shapes. These methods concatenate all features into one single feature to adapt to the hand-craft distance metric or distance metric learning setting. However, this concatenations not physically meaningful because each feature has a specific statistical property \cite{xu2013survey}. Therefore, it can not exploit the complementary geometric information to discriminate the 3D shapes well. 

Meanwhile, many researchers focus on multi-view learning methods \cite{wang2018multiview}, which makes a significant development in machine learning fields \cite{wu2018and,wu2018deepatten,wu2018andwhere}. In their mind, a real-world object may have different descriptions from multi-view observation spaces. These spaces usually look different from each other but are highly related. The multi-view setting is usually combined with single view based on either the consensus principle or the complementary principle to improve the performance of various tasks \cite{xu2013survey,zhai2012multiview,hardoon2004canonical,kumar2011co,xu2015multi}. Zhai \cite{zhai2012multiview} presented a multi-view metric learning method named Multiview Metric Learning with Global consistency and Local smoothness(MVML-GL) under a semisupervised learning setting. This method seeks a global consistent shared latent feature space firstly, and then a explicit mapping functions between the input spaces and the shared latent space can be learned via regularized locally linear regression. The process of these two steps can be solved by convex optimizations in closed form. Canonical Correlation Analysis (CCA) \cite{hardoon2004canonical} is a statistical methods correlating linear relationships between more variables. Kernel CCA(KCCA) explore the kernel function framework to extend the nonlinear processing ability. Kumar \cite{kumar2011co} proposed a co-regularized framework by advancing co-training for multi-view spectral clustering. Iterative optimization procedure is adopted to update the eigenvector one after another. Xu \cite{xu2015multi} proposed a Multi-view Intact Space Learning(MISL), which integrates the encoded complementary information in multiple views to discover a latent intact representation of the data. Intact space learning for multi-view learning provides a new multi-view representation method. It can be extened to supervised learning problems, but adding a hinge loss, or a multi-view loss to the objective. More related works survey have been proposed by \cite{wang2016iterative,wang2015robust,wang2017unsupervised,wang2017effective} which provide a more comprehensive introduction for the recent developments of multi-view learning methods on the basis of coherence with early methods. In Computer Graphic community, the multi-view means that the multiple angles projection of the 3D models. In order not to confuse the concepts, we use the multi-feature in the Computer Graphic community to replace the multi-view in machine learning community \cite{wang2014exploiting}.

Inspire by the multi-view learning methods \cite{wang2018beyond}, we develop a novel multi-feature distance metric learning algorithm in this paper, which can make fully use of the geometric information from multiple shape features. We introduce the multi-feature distance metric learning algorithm to construct a common distance metric for all features. For each feature, the distance of inner-class pair is less than a smaller threshold and that of each extra-class pair is higher than a larger threshold, respectively. Meanwhile, the algorithm minimize the distance between the Gaussian distributions of different features under different distance metrics based on KL-divergence. The two constraints are both adopted to obtain the common distance metric. Figure1. shows the pipeline of the proposed framework.

The organization of this paper is as follows. In Section 2, we provide a brief overview of previous related work of the local descriptor, shape features and metric learning algorithms. In Section 3, we present the detail of the multi-feature metric learning algorithm for non-rigid 3D shape retrieval. In Section 4, we show the results of our experiment. Section 5 concludes the paper.

\begin{figure*}
  \includegraphics[width=1\textwidth]{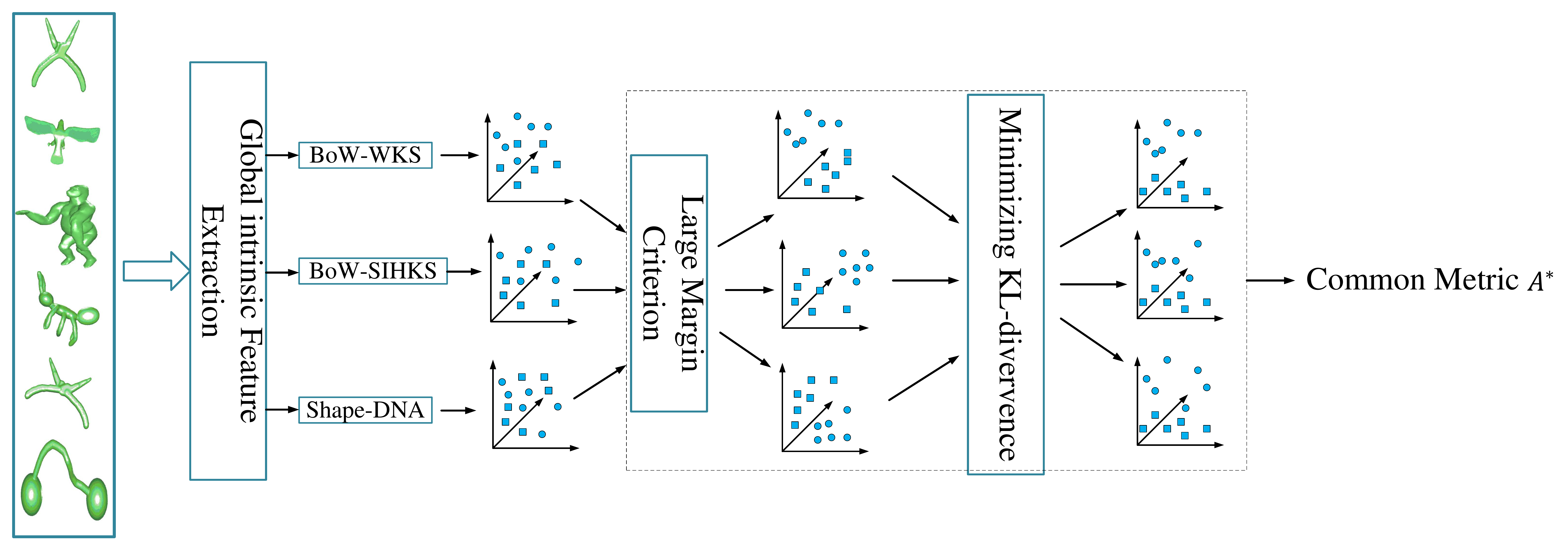}
\caption{The pipeline of the MfML based  non-rigid 3D shape retrieval}
\label{fig:1}       
\end{figure*}

\section{Related Work}
\label{sec:1}
The intrinsic feature of the shape is of importance for the non-rigid shape retrieval. Numerous works attempt to extract a discriminative and informative intrinsic feature for this task. The intrinsic feature is usually extracted by the intrinsic descriptors of the shape. Up to the present, most of the intrinsic descriptors are constructed by using spectral method, which is based on Laplace-Beltrami Operator (LBO) \cite{pinkall1993computing}. The intrinsic descriptors are often categorized as local and global. The global descriptors can be used as the feature to measure similarity among database directly. Due to the unordered of the points of the mesh, constructing an effective intrinsic global descriptors directly is difficult. The most famous intrinsic global descriptor is ShapeDNA \cite{reuter2006laplace}. It is constructed by truncating the normalized sequence of the eigenvalues of the LBO. Another effective global descriptor is based on Modal Function Transformation framework \cite{kuang2015modal}. In this framework, the spatial information of the intrinsic functions are used to construct a inner function. Then the ordered and L2 normalized eigenvalues of the inner function are adopt as the global descriptors. These two descriptors are extracted as the intrinsic features of the shapes, and the Euclidean distance or hand-crafted distance is usually used as the similarity for shape retrieval. However, the global features mainly contains the geometric structure information, and lose details of the shape. There are many point or local spectral descriptors, which contains abundant local details of the shape. Rustamove \cite{rustamov2007laplace} explore the all the spetra (eigenvalues and eigenvectors) of a shape to construct the Global Point Signature (GPS). The GPS is a intrinsic point descriptor, and robust to topological noise. But the eigenvectors are very close when the corresponding eigenvalues are close to each other. Sun \cite{sun2009concise} proposed the Heat Kernel Signature based(HKS) on heat equation. The diagonal elements of the heat kernel matrix are extracted as the HKS point descriptor. The HKS can be interpreted as the amount of heat that remains at the point of surface over a period of time. HKS is intrinsic, multi-scale and robust, which is useful for non-rigid shape analysis. However, HKS is sensitive to the change of the shape scale. Bronstein \cite{bronstein2010scale} introduced a scale-invariant version of HKS by using the Fourier transform, which moves from the time domain to the frequencies domain. Then, Aubry \cite{aubry2011wave} proposed Wave Kernel Signature based on Schrodinger equation, which describes the average probability over time to locate a particle with a certain energy distribution at the point on the surface. WKS clearly separates influences of different frequencies, treating all frequencies equally. Hence WKS reserves more high frequency information than HKS. A comprehensive survey in \cite{li2014spatially,limberger2017spectral} provides more details of the spectral signatures. 

\begin{figure*}
  \includegraphics[width=1\textwidth]{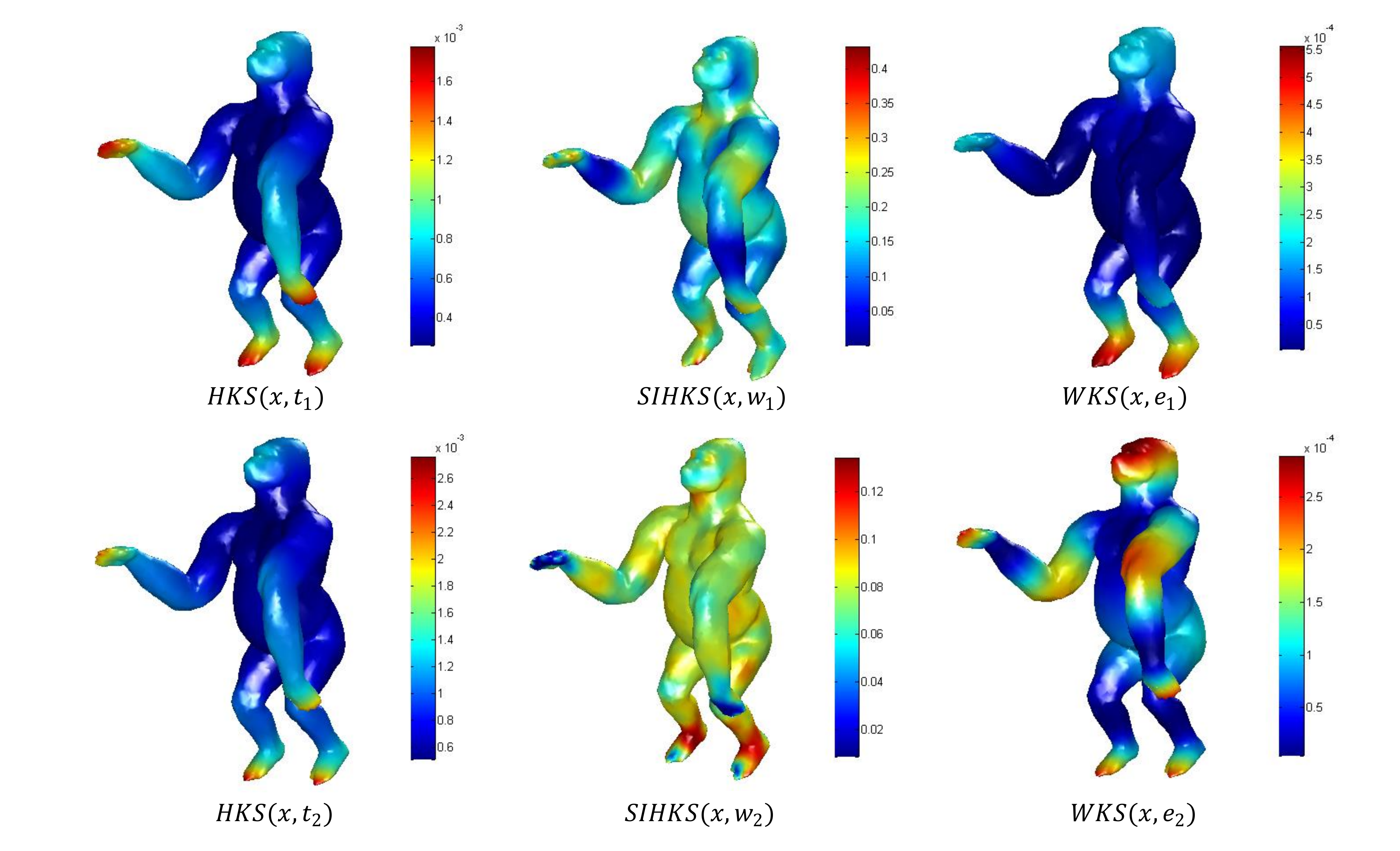}
\caption{The HKS, WKS and SIHKS point signatures with different parameters}
\label{fig:2}       
\end{figure*}

As mentioned above, although the global shape descriptors can be used for shape retrieval directly, the lack of details limit their performance on some benchmark in which the shapes contain many details. Therefore, make full use of the point or local descriptors is important for the non-rigid shape retrieval task. Many approaches aggregate the point descriptors, regions or partial views to construct the global intrinsic features by using various algorithms. Among the algorithms, Bag of Words (BoW) is the most popular one. BoW had been successfully applied to computer vision, natural language processing, etc. In recent years, it has been concerned in shape retrieval field[2]. The geometric equivalent of ‘words’ are local descriptors, which are quantized in a ‘geometric dictionary’ to obtain the ‘bag-of-geometric words’ \cite{litman2014supervised}. This algorithm codes the local descriptor to construct a global feature, in which contains rich details of the shape. Bronstein \cite{bronstein2011shape} exploited the BoW algorithm and added the spatial relations to extract the Spatially Sensitive Bags of Features (SS-BoF). The SS-BoF exhibited an excellent performance in SHREC’10 ShapeGoogle dataset benchmark. Litman \cite{litman2014supervised} explored supervised dictionary leaning with sparse coding algorithm for extracting the global feature based on point descriptors. Subsequently, the Fisher Vector (FV) and Super Vector (SV) algorithm are introduced to code the point descriptors \cite{limberger2015feature}. These two algorithms are similar to the BoW algorithm. The ‘dictionary’ is designed firstly by Gaussian Mixture Model, and then the local descriptors are coded by the Gaussian distributions. These algorithm contain multi-order information, which is more informative than BoW. Therefore, the FV and SV algorithms extract more comprehensive features for the shape. Unlike the BoF which aims to code the descriptors, Li \cite{li2013intrinsic} proposed a intrinsic spatial pyramid matching method for the retrieval task and also achieved a good performance. Furthermore, there are some approaches focus on the metric between the features more \cite{chiotellis2016non}. Chiotellis \cite{chiotellis2016non,xie2017deep}, use the weighted averaging directly on siHKS and WKS to construct the global features, and then explored the Large margin nearest neighbor algorithm to obtain the metric between the features. This method is very concise, efficient, and effective, and the result outperforms many methods in SHREC14 benchmark. The success of this approach is based on the LMNN algorithm. Therefore, the distance metric learning algorithm is also very important in the retrieval task. 

Appropriate similarities between samples can improve the performances of the retrieval system. During the past decade, several well-known distance metric learning methods are proposed for various fields \cite{davis2007information,weinberger2009distance,suykens1999least,wold1987principal,mika1999fisher,wang2016semantic}, such as ITML \cite{davis2007information}, LMNN \cite{weinberger2009distance}, SVMs \cite{suykens1999least}, PCA \cite{wold1987principal}, LDA \cite{mika1999fisher}, etc. These algorithms have been used for many computer vision and computer graphic tasks, such as classification, retrieval, correspondence, etc. These algorithms solve the problem that most features lie in a complex high-dimensional spaces where Euclidean metric is ineffective. However, most distance metric learning methods fail to integrate compatible and complementary information from multiple features to construct a distance metric. In order to explore more useful information for various applications, many researchers invest many methods to combine multi-view setting to distance metric learning algorithm. Kan \cite{kan2016multi} proposed a multi-view discriminant analysis as an extension of LDA, which has achieved excellent performances facing with multi-view features. Wu \cite{wu2016online} proposed an online multi-modal distance metric learning which has been successfully applied in image retrieval.


\section{Proposed Approach}
In this section, we introduce the proposed multi-feature metric learning algorithm (MfML) for 3D non-rigid shape retrieval in detail. We extract different types of 3D intrinsic features. Some features are global intrinsic shape descriptors, which is used to describe the global structure of the shapes. And some features are extracted by using the BoW algorithm to  code different types of point descriptors, which is used to code the geometric information of local points based on various scales. These intrinsic multiple features are used to train a common metric, which fully integrates compatible and complementary information from them. Then, we illustrate the optimization of the algorithm. 

\subsection{The Structure of Multi-feature Metric Learning}
Let $X^{v} = [x^{v}_{1},x^{v}_{2},...,x^{v}_{N}]\in R^{d_{v}\times N}, v = 1,2,...,m$ be the training set of the $v$th intrinsic feature, where $x^{v}_{i}\in R^{d_{v}}$ is $i$th samples and $N$ is the total number of samples. The Mahalanobis distance metric learning algorithm try to obtain a square matrix as the metric matrix. For $v$th features, the distance between any two samples $x^{v}_{i}$ and $x^{v}_{j}$ can be computed as:
$$d_{v}(x^{v}_{i},x^{v}_{j}) = \sqrt{(x^{v}_{i}-x^{v}_{j})^{T}A_{v}(x^{v}_{i}-x^{v}_{j})},$$
with the $A_{v}$ being decomposed as:
$$A_{v} = L_{v}^{T}L_{v}.$$
And then the $d_{v}(x^{v}_{i},x^{v}_{j})$ can also be written as:
$$d_{v}(x^{v}_{i},x^{v}_{j}) = \sqrt{(x^{v}_{i}-x^{v}_{j})^{T}A_{v}(x^{v}_{i}-x^{v}_{j})}
= \left \| L_{v}x^{v}_{i}-L_{v}x^{v}_{j} \right \|_{2}.$$
We can see from the equation that learning a Mahalanobis distance metric is equivalent to finding a linear projection onto a subspace, under which the Euclidean distance of two samples in the transformed space is equal to the Mahalanobis distance metric in the original space. We expect that the Euclidean distances between positive pairs are smaller than those between negative pairs in the subspace. Figure 2 shows the basic idea. In order to improve its discriminative ability we explore the following constraint \cite{hu2014discriminative}:
\begin{equation}
\delta_{ij}(\tau-d^{2}_{v}(x^{v}_{i},x^{v}_{j}))>1.
\end{equation}
We use $C$ to express the set that contains the pairs of samples from the same class, and $M$ to express the set that contains the pairs of samples from the different class. Let $\delta_{ij}=-1$ if $(x^{v}_{i},x^{v}_{j})\in M$ or else $\delta_{ij}=1$. Then, above constrain in equation 1 is adopted by our algorithm as follows:
\begin{equation}
   \min_{L_{v},v = 1,..,m} \sum_{v=1}^m \sum_{i,j} \frac{1}{2} g(1-\delta_{ij}(\tau-d^{2}_{v}(x^{v}_{i},x^{v}_{j})) + \sum_{v=1}^m \lambda_{v}\left \| L_{v} \right \|_{F}^2.
\end{equation}
where $g(x)=\frac{1}{\rho}log(1+\exp(\rho x))$ is a smoothed approximation of the hinge loss function, $\left \| L_{v} \right \|_{F}^2$ is the regularization term, $\lambda_{v}$ are regularization parameters. 
We can find the optimal subspace projection matrix $L_{v}, v=1,...,m$ by minimizing Eq.2. 

\begin{figure*}
  \includegraphics[width=0.8\textwidth]{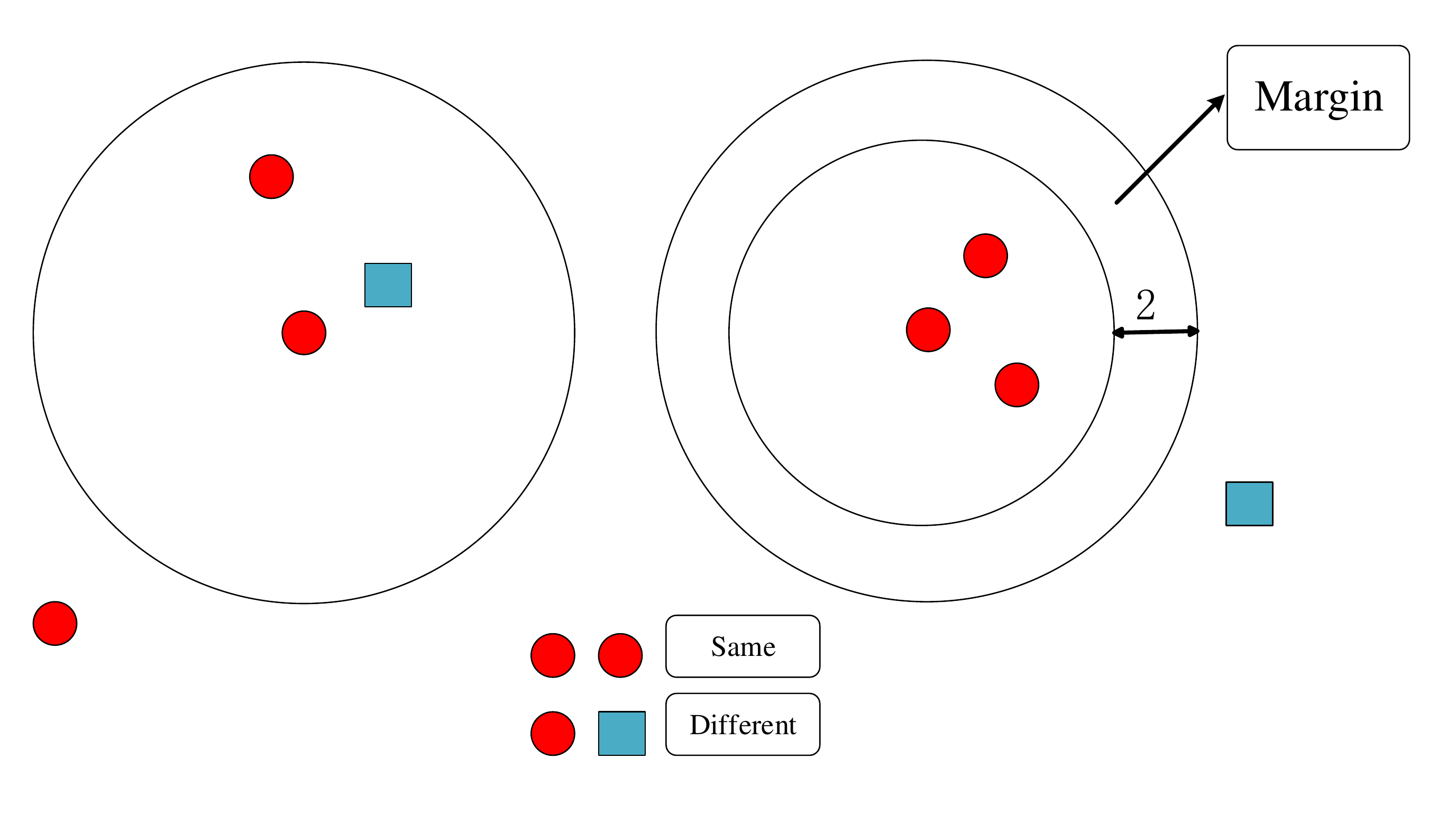}
\caption{The distance metric is optimized so that differently labeled inputs lie outside this smaller radius
by some finite margin}
\label{fig:3}       
\end{figure*}

However, it is clearly that minimizing the equation 3 equals to the sum of all features with constrain 1, which exploits neither the consensus principle nor the complementary principle for improving learning performance. Due to combine the complementary information from multiple features, we explore a hypotheses that each feature of the sample follows the Gaussian distribution with a Mahalanobis distance parameterized by $L_{v}^{T}L_{v}$, and all the distributions are similar. Inspired by ITML \cite{davis2007information} and CMSC \cite{kumar2011co}, we formulate the following cost function to measure the disagreement between the metrics $A_{v}$ and the consensus one $A_{*}$:
\begin{equation}
     \min_{L_{v}}KL(p(x^{v};A^{*})||p(x^{v};L_{v}^{T}L_{v}))
\end{equation}
where $p$ is a multivariate Gaussian as $p(x;A) = \frac{1}{Z}\exp(-\frac{1}{2}(x-\mu)^TA(x-\mu))$, and where $Z$, $\mu$ is a normalizing constant and the mean vector respectively. The $A^{*}\in R^{n \times n}$ is defined as $A^{*} = \epsilon I + \frac{1}{m}(L_{1}^{T}L_{1}+L_{2}^{T}L_{2}+...+L_{m}^{T}L_{m})$. $A^{*}$ can be treated as the common distance metric for all features. The optimization of equation 3 makes all the Gaussian distributions to be similar, which induces that every $A^{v}$ is closed with $A^{*}$. Hence, by adopting two constrains, we can formulate a new cost function to construct a new metric:
\begin{equation}
\begin{split}
    \min_{L_{v},v = 1,..,m} &\sum_{v=1}^m \sum_{i,j} \frac{1}{2} g(1-\delta_{ij}(\tau-d^{2}_{v}(x^{v}_{i},x^{v}_{j})) \\
    & + \beta \sum_{v=1}^m KL(p(x^{v};A^{*})||p(x^{v};L_{v}^{T}L_{v})) + \sum_{v=1}^m \lambda_{v}\left \| L_{v} \right \|_{F}^2 
\end{split}
\end{equation}
where $\beta$ is the parameter to balance trade-off between two constrains. We can see from the equation 4 that MfML can separate the samples from different classes by using information from multiple features. The consensus $A_{*}$ is constructed by all $A_{v}$, which fully integrates the complementary information from every feature. Meanwhile, we can see from the optimization process that the update of $A_{v}$ is also affected by $A_{*}$ .

\subsection{Optimization Process of MfML}
In this section, we provide the detail of the optimization process. Computing the gradient directly based on the definition of $KL$ divergence is difficult. Hence, we reference ITML \cite{davis2007information} to simiplify the second term as:
$$KL(p(x^{v};A^{*})||p(x^{v};L_{v}^{T}L_{v})) = \frac{1}{2} D_{\ell d}(L_{v}^{T}L_{v},A^{*})$$
where $ D_{\ell d}(L_{v}^{T}L_{v},A^{*}) = tr(L_{v}^{T}L_{v},A^{*^{-1}})-\log{\det{( L_{v}^{T}L_{v}A^{*^{-1}}})}-n$.  The $D_{\ell d}(A,B)$ is called Burg matrix divergence(or the LogDet divergence), which is a convex functions defined over matrices. And then, the cost function can be reformulated as follows:
\begin{equation}
\begin{split}
    \min_{L_{v},v = 1,..,m}
    &\sum_{v=1}^m \sum_{i,j} \frac{1}{2} g(1-\delta_{ij}(\tau-d^{2}_{v}(x^{v}_{i},x^{v}_{j})) \\
    & + \alpha\sum_{v=1}^m (tr(L_{v}^{T}L_{v},A^{*^{-1}})-\log{\det{( L_{v}^{T}L_{v}A^{*^{-1}}})}-n)\\
    & + \sum_{v=1}^m \lambda_{v}\left \| L_{v} \right \|_{F}^2 
   \end{split} 
\end{equation}

In order to solve the Eq.5, an alternating minimization is carried out. We optimize one $L_{v}$ at one time with other variables fixed by gradient descent algorithm. The consensus metric $A^{*}$ is updated after optimizing every $L_{v}$. And then, the $L_{v}$ are updated based on the new $A^{*}$. We explore the Gradient Descent (GD) to solve $L_{v}$ as:
\begin{equation}
\begin{split}
L_{v}^{t+1} = & L_{v}^{t} - \epsilon ( L^{v}_{t}\sum_{ij}\frac{\delta_{ij}(x_{i}-x_{j})(x_{i}-x_{j})^{T}}{1+\exp{(\beta z_{ij})}}+2L_{v}^{t}A^{*^{-1}}\\
& -2(L_{v}^{t^{T}}L^{t}_{v})^{\dagger}L^{t}_{v}+2\lambda_{v}L^{t}_{v})
\end{split}
\end{equation}
where $z_{ij} = 1-\delta_{ij}(\tau-d^{2}_{v}(x^{v}_{i},x^{v}_{j}))$.
At last, we can a consensus metric matrix $A^{*}$ as the output of the MfML algorithm. The $A^{*}$ can be  directly used for measuring the similarity between the any type features that have been preprocessed by PCA for unifying the dimension. From the procedure of updating $L_{v}$ and $A^{*}$, we can see that the information from multiple feature is integrated into a co-regularized framework. 

\section{Experiment}
In this section, we demonstrate the results of non-rigid 3D shape retrieval based on MfML, and then compare it with the state-of-the-art non-rigid 3D shape retrieval approaches on SHREC'11 \cite{lian2011shape} and SHREC'15 \cite{lian2015shrec,lian2010shrec} benchmark dataset. The experiment is conducted on a 3.0 GHz Core(TM) i7 computer with 16GB memory.  
\subsection{Experiment Setting}
For all 3D shape benchmark datasets, we explore 2 different types of point signatures and 1 global descriptor to form multiple shape features. We show the setting of the point signatures and the global descriptor used in our experiment as follows:

1)WKS: The Wave Kernel signature describes the average probability over time to locate a particle with a certain energy distribution at the point on the surface \cite{aubry2011wave}. WKS clearly separates influences of different frequencies, treating all frequencies equally, and organizes the intrinsic geometric information of the point in a multi-scale way. 

2)siHKS: The scale-invariant Heat Kernel Signature (siHKS) is a scale-invariant version of heat kernel descriptor \cite{bronstein2010scale}. The construction is based on a logarithmically sampled scale-space, and then the absolute values of  Fourier transform are used for moving the scale factor from the time domain to the frequencies domain.

3)ShapeDNA: The ShapeDNA is constructed by truncating the normalized sequence of the eigenvalues of the LBO \cite{reuter2006laplace}. The main advantages of ShapeDNA are the simple representation, comparison, scale invariance. And in spite of its simplicity, it has a good performance for non-rigid shape retrieval.

We use the first 100 eigenvectors of LBO to construct two point signatures. The 100-dimensional WKS with setting the variance to 6 and 50-dimensional siHKS with same setting as in \cite{litman2014supervised} are extracted by them. Then we explore the BoW algorithm to code the WKS and siHKS respectively, and then we can obtain the 64-dimensional BoW-WKS and BoW-siHKS global features. We utilize the first 40 normalized eigenvalues of the LBO  as the ShapeDNA feature. PCA is used to project all features into a 30 dimension subspace as the pre-processing of our experiment.

\subsection{Experiment on SHREC'11}
In this section, we conduct 2 experiments on SHREC'11 benchmark dataset. The database contains 600 watertight meshes, which is derived from 30 original models. Every class contains 1 null model and 19 deformed models based on it. Firstly we compare method based on MfML with the methods related with LBO: (1)ShapeGoogle \cite{bronstein2011shape}, 2)Modal Function Transformation(MFT) \cite{kuang2015modal}, 3)Supervised Dictionary Learning(SupDL) \cite{litman2014supervised}, and these three features without being integrated by MfML. We randomly select 60\% samples with the labels from every class as the training set. In test stage, we project all features into a 30-dimensional subspace, and explore the MfML to calculate the common metric $A^{*}$. We compare with 1).ShapeDNA, 2)BoW-WKS, 3)BoW-siHKS, 4)ShapeGoogle, 5)MFT, 6)SupDL. The test set are disjoint with the training set. The PR(precision-recall)-curves show in fig. Next experiment, the test is taken for all the dataset. We compare our MfML approach with the method in \cite{lian2011shape}: FVF-WKS, BOW-LSD, LSF, SD-GDM, FOG, MDS-CM-BOF, and MeshSIFT. We evaluate the retrieval performance based on the quantitative measures from PSB \cite{shilane2004princeton}: Nearest Neighbor (NN), First Tier (FT), Second Tier (ST), Emeasure (E), and Discounted Cumulative Gain (DCG) \ref{tab:1} . The results are averaged over 5 runs with different training set. 

\begin{figure*}
  \includegraphics[width=0.7\textwidth]{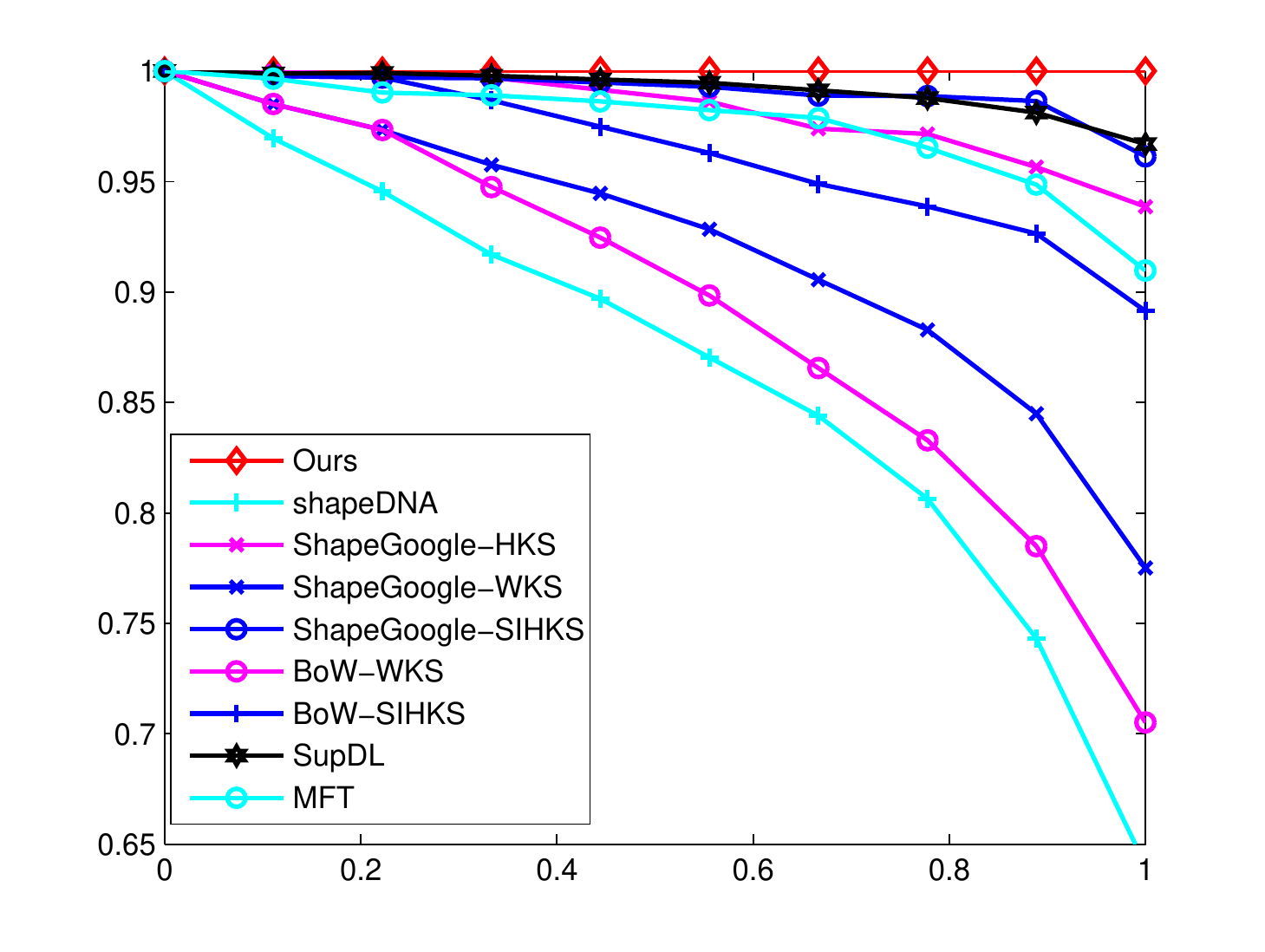}
\caption{Comparison of the precision Recall curves (PR-curves)
among our method and the other methods on SHREC’11 Non-rigid
dataset. }
\label{fig:4}       
\end{figure*}

\begin{table}
	\centering
	\caption{Five quantitative measures on SHREC'11}
	\label{tab:1}
	\begin{tabular}{lccccc}
		\hline
		Method &NN &FT &ST &E &DCG \\ \hline		
		FOG &96.8 &81.7 &90.3 &66.0 &94.4\\
		\hline
		BOW-LSD &95.5 &67.2 &80.3 &57.9 &89.7\\
		\hline
		MDS-CM-BOF &99.5 &91.3 &96.9 &71.1 &98.2\\
		\hline
		LSF &99.5 &79.9 &86.3 &63.3 &94.3\\
		\hline
		SD-GDM &100 &96.2 &98.4 &73.1 &99.4 \\
		\hline
		MeshSIFT &99.5 &88.4 &96.2 &70.8 &98.0\\ 
		\hline		
		\textbf {Our method} & \textbf{100} & \textbf{100} & \textbf{100} & \textbf{74.5}  & \textbf{100}\\ \hline	
	\end{tabular}
	
\end{table}

\subsection{Experiment on SHREC'15}
In this section, we conduct 2 experiments on SHREC'15 benchmark dataset also. The database contains 1200 watertight meshes, which is derived from 50 original models. Every class contains 1 null model and 23 deformed models based on it. This dataset contains all the models in SHREC'11 dataset. In every class, 20 models have the same topological structures as the original model, and topological structures of other 4 objects are modified by parts being connected, which is more challenging.  We randomly select 70\% samples with the labels from every class as the training set. In test stage, we use PCA to project all features into a 30-dimensional subspace, and the MfML to calculate the common metric $A^{*}$. We compare with 1).ShapeDNA, 2)BoW-WKS, 3)BoW-siHKS, 4)ShapeGoogle, 5)MFT, 6)SupDL. The test set are disjoint with the training set. The PR-curves show in fig. Next experiment, the test is taken for all the dataset. We compare our MfML approach with the method in \cite{lian2015shrec,lian2010shrec}: HAPT, SG\_L1, FVF-WKS, SID, and EDBVF\_NW  \ref{tab:2}. The results are averaged over 10 runs with different training set.
\begin{figure*}
  \includegraphics[width=0.7\textwidth]{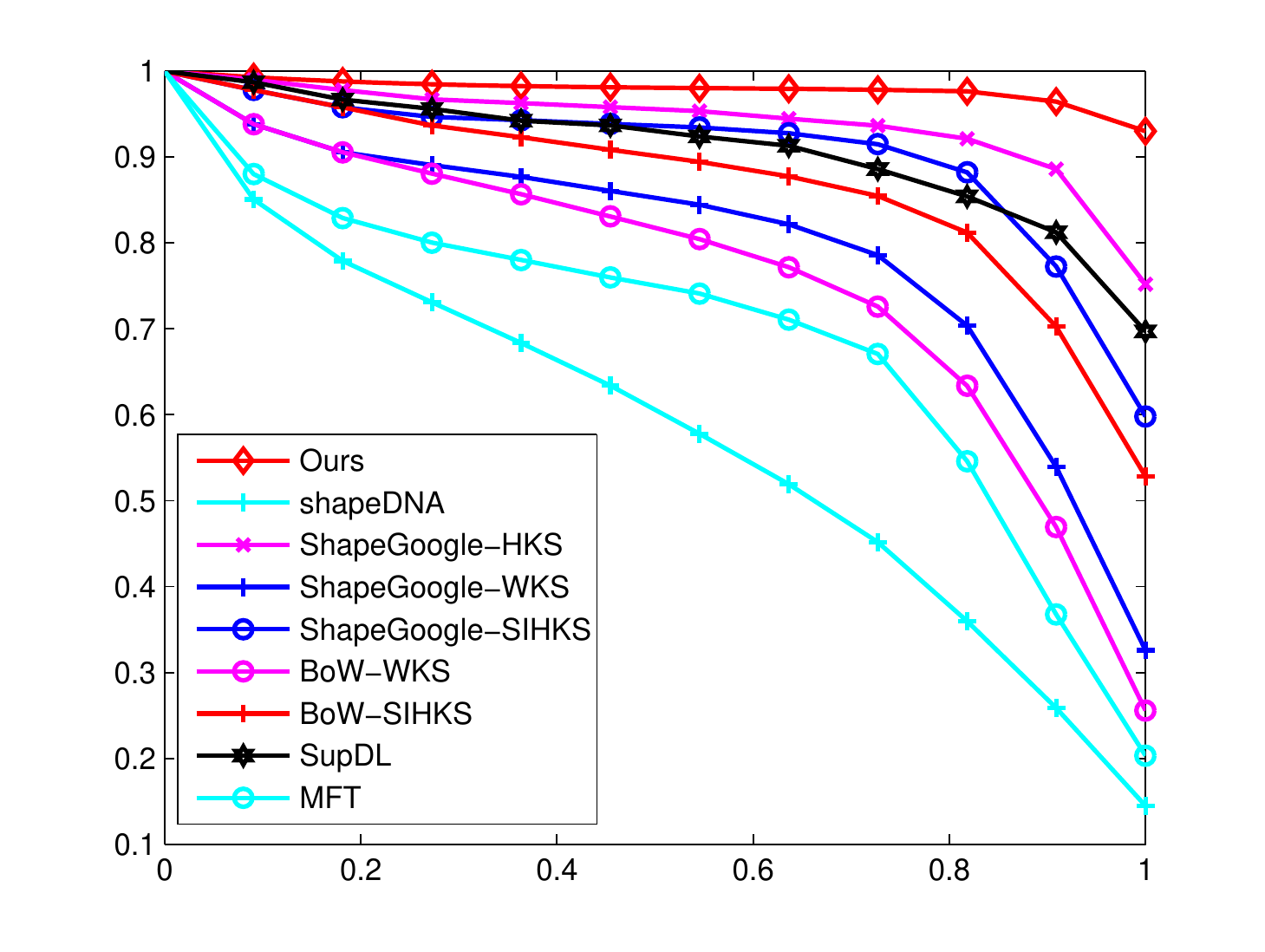}
\caption{Comparison of the precision Recall curves (PR-curves)
among our method and the other methods on SHREC’15 Non-rigid
dataset. }
\label{fig:5}       
\end{figure*}

\begin{table}
	\centering
	\caption{Five quantitative measures on SHREC'15}
	\label{tab:2}
	\begin{tabular}{lccccc}
		\hline
		Method &NN &FT &ST &E &DCG \\ \hline			
		HAPT &\textbf{100.0} &96.1 &97.9 &81.2 &\textbf{99.9}\\
		\hline
		SG\_L1 &97.3 &75.9 &81.4 &65.9 &91.9\\
		\hline
		FVF-WKS &100 &82.5 &86.3 &88.3 &71.8\\
		\hline
		SID &97.7 &71.9 &82.1 &64.8 &92.0 \\
		\hline
		EDBCF\_NW &97.8 &79.3 &83.4 &70.8 &94.3\\ \hline
		\textbf{Our method} & 100 & \textbf{99.2} & \textbf{99.7} & \textbf{82.7}  & 99.5\\ \hline				
	\end{tabular}	
	
\end{table}

\subsection{Experiment Result}
We can clearly find from fig and fig that MfML outperforms other methods based on LBO and the features without MfML. Specially, MfML perfectly discriminate all types of models in SHREC'11. Meanwhile, in SHREC'15 we have the best performance, in which the  precision is close to 1. The comparison with the state-of-the-art methods in \cite{lian2011shape} are demonstrated in table and table. The MfML outperforms other methods in SHREC'11. Meanwhile, HAPT can outperform the MfML for quantitative measures in SHREC'15. Even though FVF-WKS can achieve better performance in some quantitative measures, MfML is a better method for more datasets. 

\section{Conclusion}
In this paper, we proposed a novel multi-feature metric learning method for non-rigid 3D shape retrieval. MfML aims to exploit compatible and complementary geometric information from multiple intrinsic features. For each feature, MfML makes the distance of inner-class pair less than a smaller threshold and that of each extra-class pair higher than a larger threshold, respectively. Meanwhile, by minimizing KL-divergence between the Gaussian distributions of different features under different distance metrics to let multiple features to work together to obtain a consensus distance metric. The two constraints are both adopted to obtain an excellent common distance metric. Many experiments on two benchmark datasets have verified that MfML is a highly efficient multi-feature distance metric learning method.

\section*{Acknowledge}
This study was funded by the National Natural Science Foundation of China Grant 61370142 and Grant 61272368, by the Fundamental Research Funds for the Central Universities Grant 3132016352, by the Fundamental Research of Ministry of Transport of P.R. China Grant 2015329225300. Huibing Wang, Haohao Li and Xianping Fu declare that they have no conflict of interest. Huibing Wang and Haohao Li contribute equally to this article. This article does not contain any studies with human participants or animals performed by any of the authors.

\bibliographystyle{unsrt}
\bibliography{citepaper}   


\end{document}